\documentclass[a4paper,english]{IEEEtran}
\usepackage[T1]{fontenc}
\usepackage[latin9]{inputenc}
\usepackage{float}
\usepackage{amsmath}
\usepackage{graphicx}
\usepackage{rotfloat}

\makeatletter

\pdfpageheight\paperheight
\pdfpagewidth\paperwidth

\makeatother

\usepackage{babel}
\begin{document}
\title{Comparison of Embedded Spaces for \\
 Deep Learning Classification}
\author{Stefan Scholl\\
research@panoradio-sdr.de}
\maketitle
\begin{abstract}
Embedded spaces are a key feature in deep learning. Good embedded
spaces represent the data well to support classification and advanced
techniques such as open-set recognition, few-short learning and explainability.
This paper presents a compact overview of different techniques to
design embedded spaces for classification. It compares different loss
functions and constraints on the network parameters with respect to
the achievable geometric structure of the embedded space. The techniques
are demonstrated with two- and three-dimensional embeddings for the
MNIST, Fashion MNIST and CIFAR-10 datasets, allowing visual inspection
of the embedded spaces. 
\end{abstract}

\begin{IEEEkeywords}
Deep Learning, Embedded Space, Metric Learning, Softmax Loss, Center
Loss, Angular Margin Loss, Contrastive Loss, Triplet Loss
\end{IEEEkeywords}

\section{Introduction to Embedded Spaces for Classification}

An embedded space is a lower-dimensional representation of high-dimensional
input data that is learned by a neural network. The embedding preserves
relevant properties and relationships of the original data and helps
to capture semantic similarities and differences within the data.
Typical embeddings, for example, group similar data geometrically,
while dissimilar data is located further apart.

Embedded spaces are inherently created when training deep neural networks,
e.g. for classification with softmax loss. However, there are also
advanced methods, often referred to as \emph{metric learning}, which
can generate more expressive embeddings. Good embeddings support classification
and advanced techniques like open-set recognition \cite{A.Mahdavi.2021}
and few-shot learning \cite{ArchitParnami.2022}. Moreover, they can
contribute to explainability and interpretability of neural networks.

\begin{figure}
\begin{centering}
\includegraphics[scale=0.6]{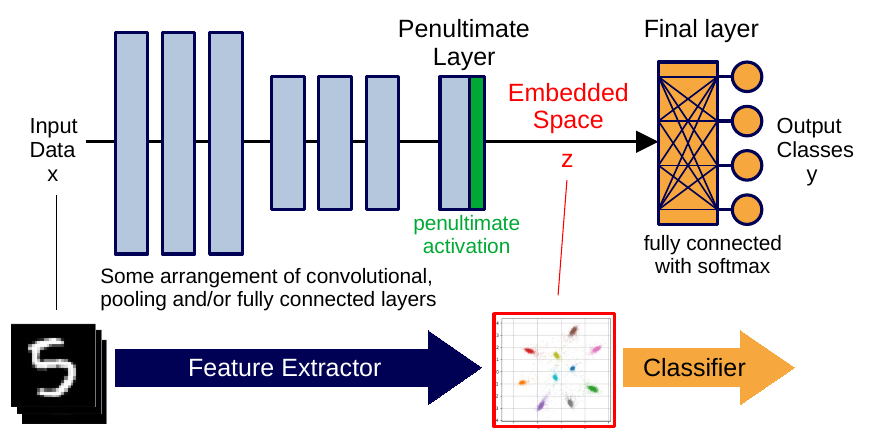}
\par\end{centering}
\caption{\label{fig:Typical-convolutional-network}Typical convolutional network
for classification with the embedded space as it is considered in
this paper}

\end{figure}

The structure of a typical deep neural network for classification
and its embedded space is shown in Figure \ref{fig:Typical-convolutional-network}.
The neural network can be divided into two parts: the \emph{feature
extractor} and the \emph{classifier}, with the embedded space located
in between. The feature extractor provides a mapping from the input
space to the embedded space (or feature space) and covers all layers
from the first to the penultimate layer. It can contain multiple convolutional,
pooling and fully connected layers. Following the feature extractor,
the classifier calculates the class scores from the embedded space.
It typically consists of a fully connected linear layer followed by
a softmax operation.

The embedded space in classification tasks is unfortunately sometimes
neglected and left to the softmax-based optimization process alone.
However, the designer of a neural network can influence the embedded
space to create a more useful and meaningful structure.

This paper summarizes popular techniques to improve the geometry of
the embedded space in order to obtain a more regular structure. It
includes additional network constraints and advanced loss functions,
such as center loss, angular margin losses, contrastive and triplet
loss. This additional effort can support the interpretability and
explainability of the neural network, as well as open-set recognition
and few-shot learning. An exemplary study on the MNIST, Fashion MNIST
and CIFAR-10 datasets allows for visual inspection and comparison
of the different approaches.

\section{Design Methods for Deep Embedded Spaces}

\subsection{Overview}

The structure of the embedded space is influenced by two factors:
\begin{enumerate}
\item The loss function
\item Constraints on the network components, such as weights, biases or
activation functions
\end{enumerate}
By taking care of these factors, the designer of a neural network
can influence the geometry of the embedded space. 

In the following, different techniques for shaping the embedded space
are presented in detail. The results are demonstrated with the relatively
simple MNIST dataset, which allows the use of a two-dimensional embedded
space that is easy to visualize and inspect. However, it should be
noted that for many other datasets more than two dimensions are required
to obtain good embeddings. The example neural network for feature
extraction in this paper is a simple CNN with Conv(32) -> Pool ->
Conv (64) -> Pool -> Conv (128) -> Fully Connected (256) -> Fully
Connected (2).

\subsection{Penultimate Layer Activation Function}

\begin{figure}
\begin{centering}
\includegraphics[scale=0.67]{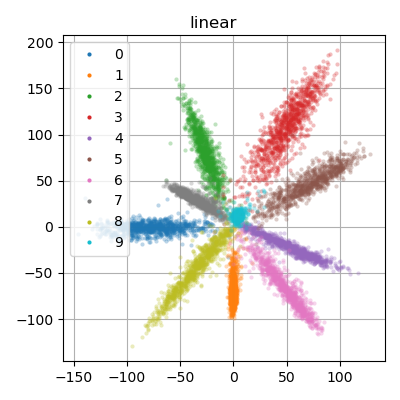}\includegraphics[scale=0.4]{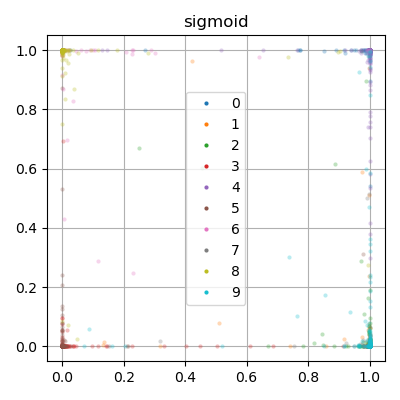}
\par\end{centering}
\begin{centering}
\includegraphics[scale=0.4]{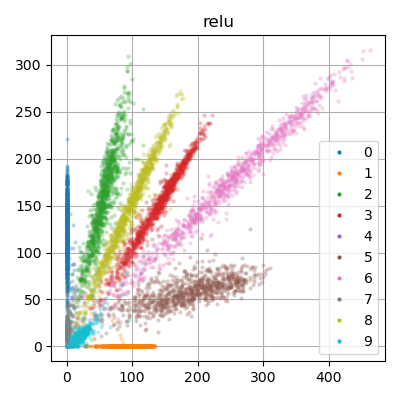}\includegraphics[scale=0.4]{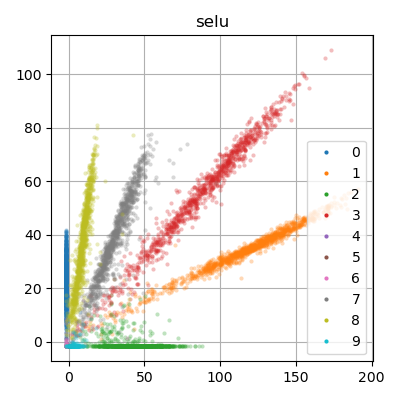}
\par\end{centering}
\caption{\label{fig:penultimate-activation}Comparison of different activation
functions (linear, sigmoid, relu, selu \cite{Geron.October2022})
of the penultimate layer, i.e. right before the embedding (loss here
is softmax loss). The linear activation provides the most suitable
structure. (MNIST dataset)}

\end{figure}

The penultimate layer is the layer directly in front of the embedded
space (see Figure \ref{fig:Typical-convolutional-network}). Therefore,
its activation function has a large influence on the embedding. Figure
\ref{fig:penultimate-activation} shows the embedded space for different
common activation functions. It is easy to see that a linear activation
function provides the most visually interpretable and information
rich structure. Therefore, in this paper a linear penultimate activation
is used as the basis for all following techniques.

\subsection{Softmax Loss}

The typical approach for classification is to use the softmax activation
for the last layer and to combine it with the categorical cross-entropy
loss \cite{Geron.October2022}. Mathematically, this combination can
be described by the softmax loss \cite{Liu.2017}:

\[
L_{softmax}=-log\frac{e^{W_{y}^{T}z+b_{y}}}{\sum_{j=1}^{M}e^{W_{j}^{T}z+b_{j}}}
\]

where $z$ is the feature vector in the embedded space of an input
data sample $x$ after propagation through the feature extractor neural
network. $M$ is the number of classes and $y$ is the ground truth
label of the input data $x$. $W_{j}$ and $b_{j}$ are the weights
and biases of the final linear classification layer for class $j$.
Note that for simplicity of the notation, the loss functions are provided
for a single data sample rather than a complete batch throughout this
article.

The softmax loss mainly optimizes the classification accuracy and
does not introduce any strong geometric conditions on the embedded
space.

\begin{figure}
\begin{centering}
\includegraphics[scale=0.43]{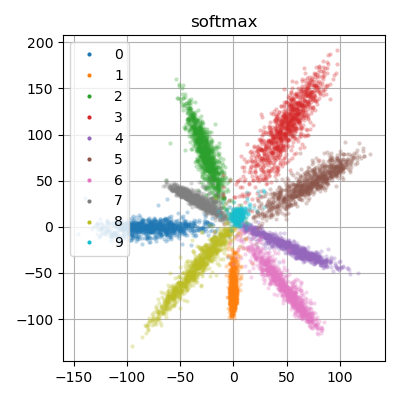}\includegraphics[scale=0.43]{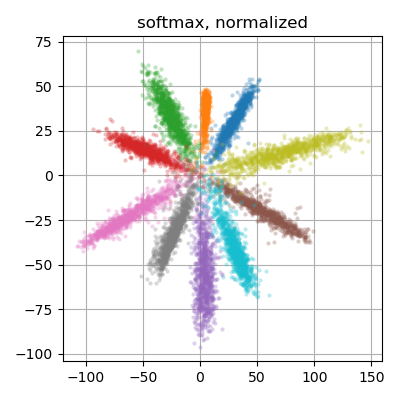}
\par\end{centering}
\caption{\label{fig:softmax}Embedded space for MNIST with ordinary softmax
loss (left) and softmax loss with normalized weights for a more regular
geometry (right)}

\end{figure}

An additional normalization step can introduce a more regular structure
to the embedded space. It sets the bias values to zero and forces
the magnitude of the weight vectors to be 1, i.e \\
\[
\left\Vert W_{j}\right\Vert =1,\;b_{j}=0
\]
This additional restrictions usually do not impact the overall classification
accuracy. Embedded spaces for softmax with and without normalization
are shown in Figure \ref{fig:softmax}.

\subsection{Angular Margin Loss}

Angular margin losses are a family of advanced loss functions that
optimize the angles between classes in the embedded space, while introducing
an additional margin to enforce more compact classes.

Different angular margin losses have been proposed in literature,
which are based on a reformulated softmax loss. The softmax loss depends
on the dot product between the weight vector $W_{j}$ and the data
point in the embedded space $z$, which can be rewritten as
\[
W_{j}^{T}z+b_{j}=\left\Vert W_{j}\right\Vert \left\Vert z\right\Vert cos(\theta_{j})+b_{j}
\]
where $\theta$ denotes the angle between $W_{j}$ and $z$. Angular
losses typically assume normalization:

\[
\left\Vert W_{j}\right\Vert =1,\;\left\Vert z\right\Vert =1,\;b_{j}=0.
\]
 Then the re-formulated softmax loss is

\begin{align*}
L_{softmax} & =-log\frac{e^{W_{y}^{T}z}}{\sum_{j=1}^{M}e^{W_{j}^{T}z}}=-log\frac{e^{cos(\theta_{y})}}{\sum_{j=1}^{M}e^{cos(\theta_{j})}}
\end{align*}

In order to improve the compactness of the classes, an additional
margin parameter $m$ is introduced. There are different variants
of angular margin losses, that differ in the way the margin parameter
is introduced. Early versions such as L-softmax or sphereface loss
\cite{Liu.2017} suffer from shortcomings such as unevenly distributed
margins for different classes. Advanced angular margin losses such
as arcface \cite{Deng.2019} or cosface \cite{Wang.2018} solve these
drawbacks and are therefore more suitable. Here the cosface loss is
used, which is defined by

\[
L_{cosface}=-log\frac{e^{s\left(cos(\theta_{y})-m\right)}}{e^{s\left(cos(\theta_{y})-m\right)}+\sum_{j\neq y}e^{s\:cos(\theta_{j})}}
\]

where $s$ is an additional scaling parameter, see \cite{Wang.2018}.

\begin{figure}
\begin{centering}
\includegraphics[scale=0.4]{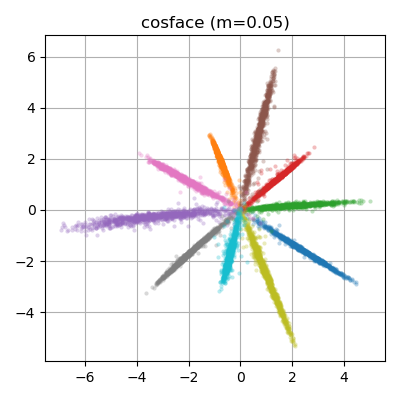}\includegraphics[scale=0.4]{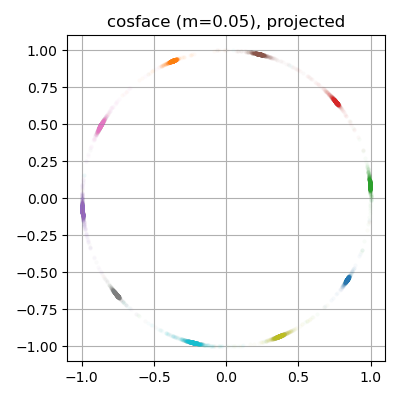}
\par\end{centering}
\begin{centering}
\includegraphics[scale=0.4]{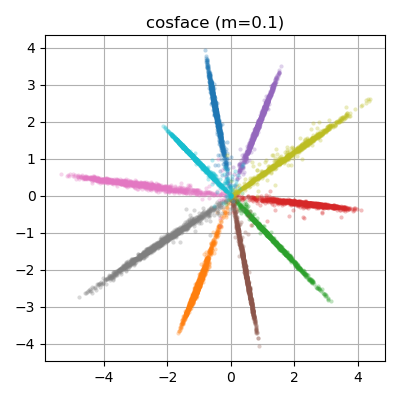}\includegraphics[scale=0.4]{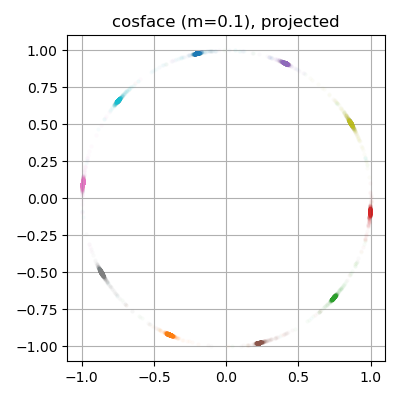}
\par\end{centering}
\begin{centering}
\includegraphics[scale=0.4]{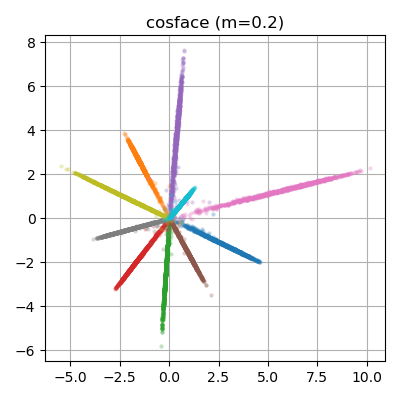}\includegraphics[scale=0.4]{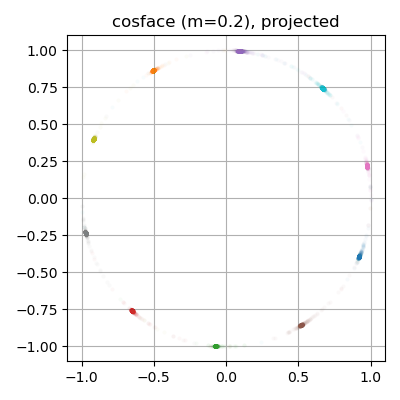}
\par\end{centering}
\caption{\label{fig:Angular-margin-loss}Angular margin loss (cosface) for
different margins for $m=0,0.05$ and $0.2$. The right column shows
the projection of the features to the unit circle. (MNIST)}
\end{figure}

In a variant of the angular margin loss, the data points in the embedded
space are projected onto a hypersphere, which removes any magnitude
information and focuses only on the angular information. Example plots
for the cosface loss with different margins are shown in Figure \ref{fig:Angular-margin-loss}.

\subsection{Center Loss}

The center loss extends the softmax loss with an additional term,
that reduces the intra-class spread \cite{Wen.2016}. This results
in visually more compact classes.

For this purpose, the additional loss term aims to minimize the Euclidean
distance between points of the same class. The complete center loss
function is given by

\begin{eqnarray*}
L & = & L_{softmax}+L_{center}\\
 & = & L_{softmax}+\frac{\lambda}{2}\left\Vert z-c_{y}\right\Vert _{2}^{2}
\end{eqnarray*}

where $c_{y}$ denotes the geometric center of class $y$, which is
the average of the points of that class in the embedded space. The
scalar value $\lambda$ weights the center loss term and thus balances
the overall loss function between the softmax and the center component.

\begin{figure}[H]
\begin{centering}
\includegraphics[scale=0.43]{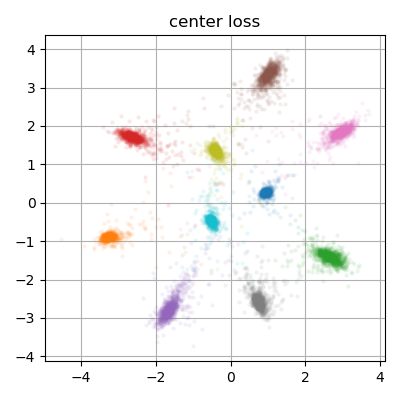}\includegraphics[scale=0.43]{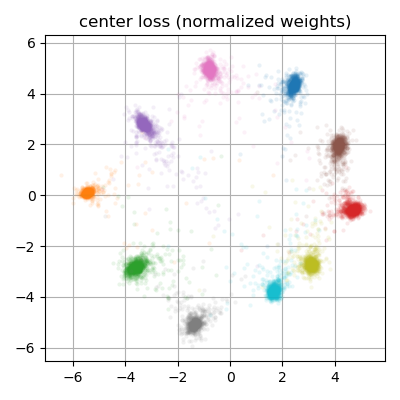}
\par\end{centering}
\caption{\label{fig:Center-loss}Center loss for an MNIST embedding without
(left) and with (right) weight normalization. The classes are more
compact than with ordinary softmax loss (Figure \ref{fig:softmax})}
\end{figure}

As with ordinary softmax, a normalization of W and bias can be introduced,
resulting in a more symmetric geometry in the embedded space. Exemplary
results for the center loss approach are shown in Figure \ref{fig:Center-loss}.

\subsection{Contrastive and Triplet Loss}

Contrastive and triplet loss operate directly on the embedded space
without the final classification layer and softmax. They basically
optimize the (Euclidean) distance between data points: Data points
of the same class are pushed closer together, while data points of
different classes are pushed farther apart. 

Contrastive loss \cite{Hadsell.2006} considers the training data
samples pairwise. If a pair is of the same class the distance between
the points is decreased. If the pair is of different classes the distance
is increased. Contrastive loss optimizes 

\[
L_{c}=\frac{1}{2}(1-Y)\left\Vert z_{0}-z_{1}\right\Vert _{2}^{2}+\frac{1}{2}Y\:max\left(0,m-\left\Vert z_{0}-z_{1}\right\Vert _{2}\right)^{2}
\]

where $Y$ indicates if the two data samples $z_{0}$ and $z_{1}$
are from the same class ($Y=0$) or from a different class ($Y=1$)
and $m$ defines a margin.

Triplet loss is closely related to the contrastive loss approach.
However, instead of working on pairs of data samples, it uses triplets,
i.e. groups of three data samples. The first sample selected is called
the anchor $z_{a}$. The second sample $z_{p}$ is from the same class
as the anchor (positive), while the third sample $z_{n}$ is from
a different class (negative). In each training step the feature extractor
network is optimized to reduce the distance between the anchor and
the positive sample and to increase the distance between the anchor
and the negative sample. The triplet loss function for Euclidean distances
is defined as

\[
L_{triplet}=max\left(\left\Vert z_{a}-z_{p}\right\Vert _{2}^{2}-\left\Vert z_{a}-z_{n}\right\Vert _{2}^{2}+m,0\right)
\]

\begin{figure}
\begin{centering}
\includegraphics[scale=0.43]{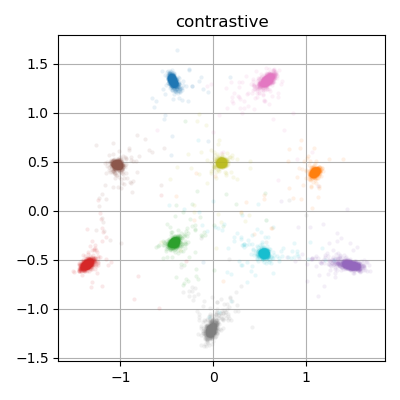}\includegraphics[scale=0.43]{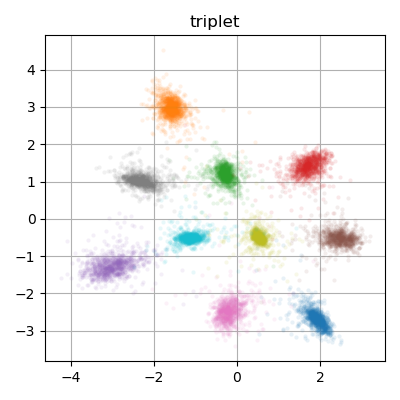}
\par\end{centering}
\caption{\label{fig:Contrastive-triplet-loss}Contrastive and triplet loss
for MNIST}
\end{figure}

Triplet loss \cite{Schroff.2015} is sensitive to the way the triplets
are selected from the training dataset, which is often called \emph{mining}.
Figure \ref{fig:Contrastive-triplet-loss} shows the embedded space
for the contrastive and the triplet loss.

\subsection{Regression}

Regression is an approach to design the embedded space manually. The
regression loss function is applied directly to the embedded space
without a classification layer, similar to the approaches of contrastive
and triplet loss.

In regression, the neural network designer first defines a target
coordinate $t_{j}$ in the embedded space for each class $j$ prior
to training. As a result, there is a one-to-one correspondence between
the target coordinates and the classes. Then a regression loss, such
as mean square error is used to train the neural network to map the
input data to the assigned target points in the embedded space. The
regression loss used here is simply given by
\[
L_{regression}=\left\Vert z-t_{y}\right\Vert _{2}^{2}
\]

where $t_{y}$ is the user-defined target point in the embedded space
for the ground truth class $y$. To obtain a classification result
during inference, the class with the smallest Euclidean distance to
the data point $z$ is output.

\begin{figure}[H]
\begin{centering}
\includegraphics[scale=0.43]{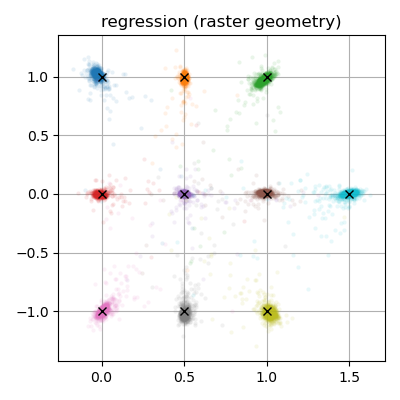}\includegraphics[scale=0.43]{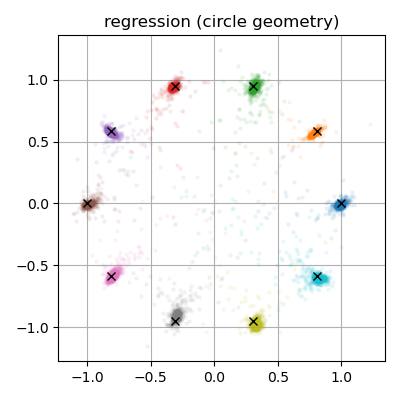}
\par\end{centering}
\caption{\label{fig:regression}Different embedded spaces designed by defined
target points (marked as black crosses) and regression loss (MNIST)}
\end{figure}

Since the designer needs to manually select the target coordinates,
arbitrary geometries are possible. While this is easy to design for
low-dimensional embeddings with two or three dimensions, it may be
more difficult when the dimensionality is higher. An open question
is also how to arrange the classes, so that the geometry of the embedded
space properly reflects the structure and semantics of the data. Figure
\ref{fig:regression} shows two variants of the target coordinates
in the two-dimensional space: a raster and a circular arrangement.

\section{Comparison of Different Embeddings}

\begin{sidewaysfigure*}
\hspace{0.4cm}\includegraphics[scale=0.27]{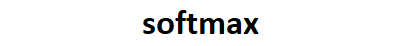}\includegraphics[scale=0.27]{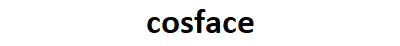}\includegraphics[scale=0.27]{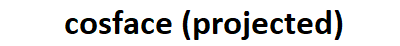}\includegraphics[scale=0.27]{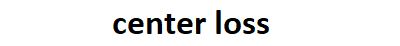}\includegraphics[scale=0.27]{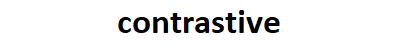}\includegraphics[scale=0.27]{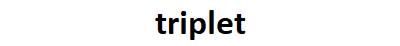}\includegraphics[scale=0.27]{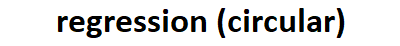}\includegraphics[scale=0.27]{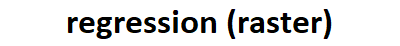}
\begin{raggedright}
\includegraphics[angle=90,scale=0.27]{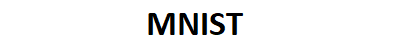}\includegraphics[scale=0.28]{bilder/softmax_linear_nn_mod_51}\includegraphics[scale=0.28]{bilder/angular_margin_cosface_02}\includegraphics[scale=0.28]{bilder/angular_margin_cosface_02_projected}\includegraphics[scale=0.28]{bilder/softmax_center3_lambda03}\includegraphics[scale=0.28]{bilder/contrastive}\includegraphics[scale=0.28]{bilder/triplet}\includegraphics[scale=0.28]{bilder/regression_circle}\includegraphics[scale=0.28]{bilder/regression_raster}
\par\end{raggedright}
\begin{raggedright}
\includegraphics[angle=90,scale=0.27]{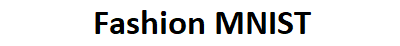}\includegraphics[scale=0.28]{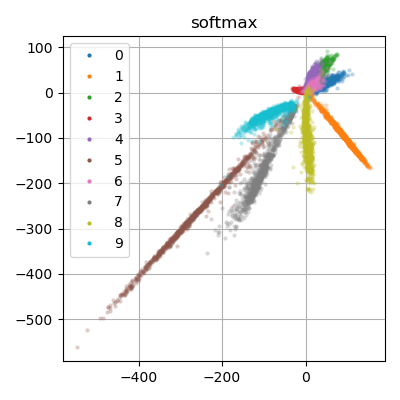}\includegraphics[scale=0.28]{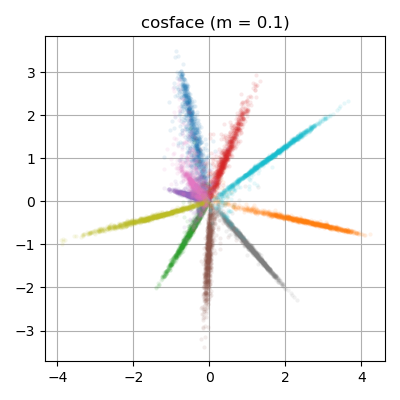}\includegraphics[scale=0.28]{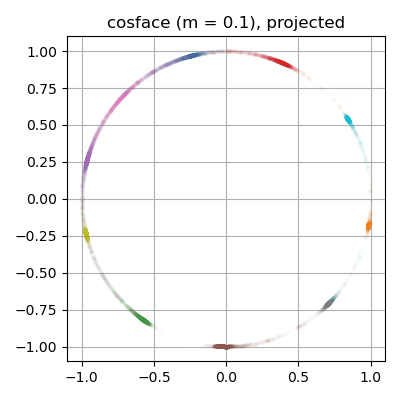}\includegraphics[scale=0.28]{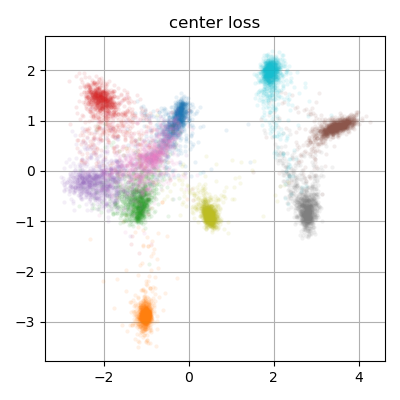}\includegraphics[scale=0.28]{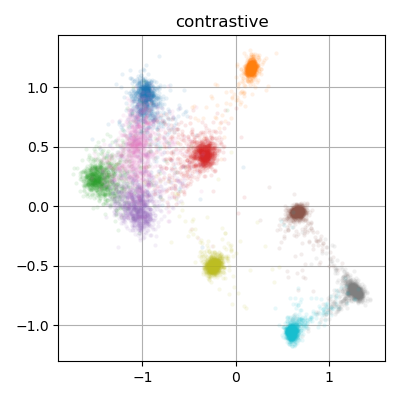}\includegraphics[scale=0.28]{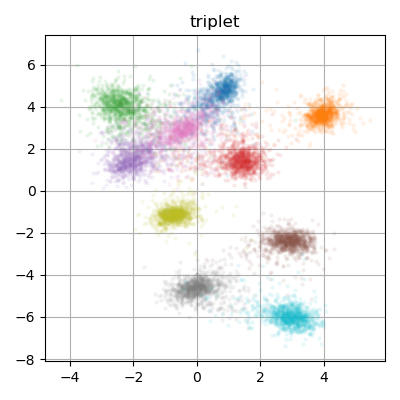}\includegraphics[scale=0.28]{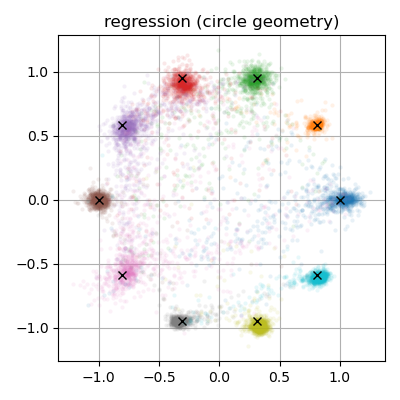}\includegraphics[scale=0.28]{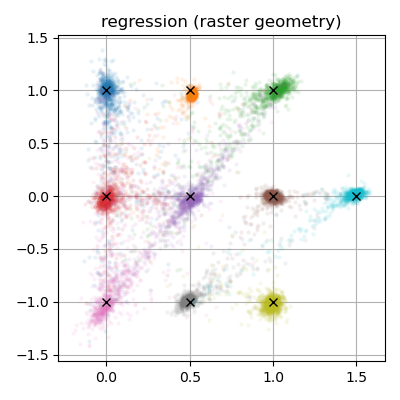}
\par\end{raggedright}
\begin{raggedright}
\includegraphics[angle=90,scale=0.27]{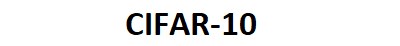}\includegraphics[scale=0.28]{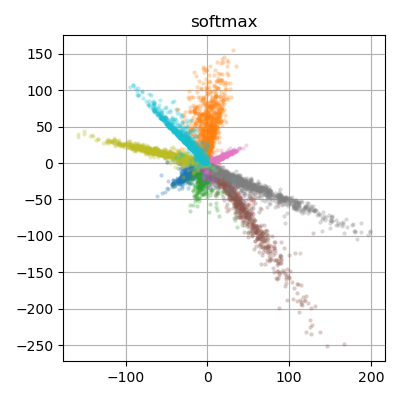}\includegraphics[scale=0.28]{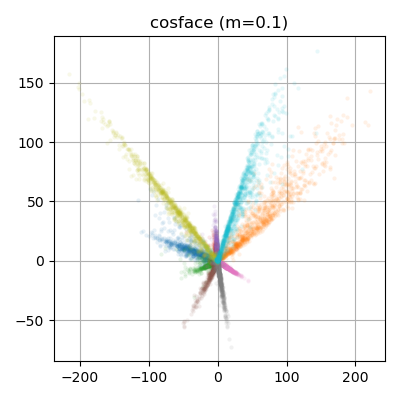}\includegraphics[scale=0.28]{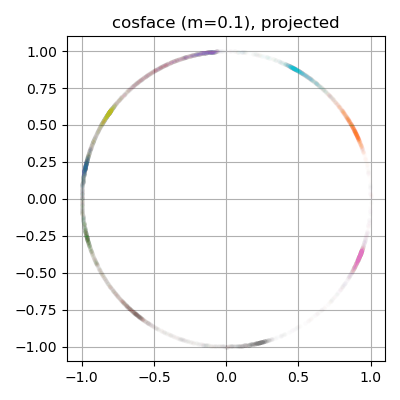}\includegraphics[scale=0.28]{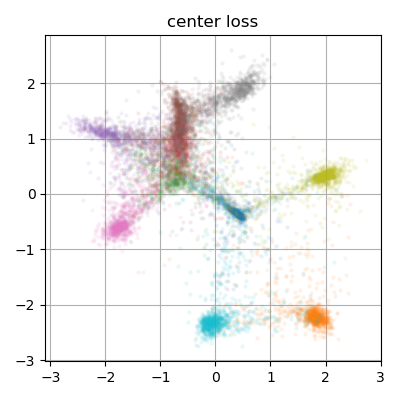}\includegraphics[scale=0.28]{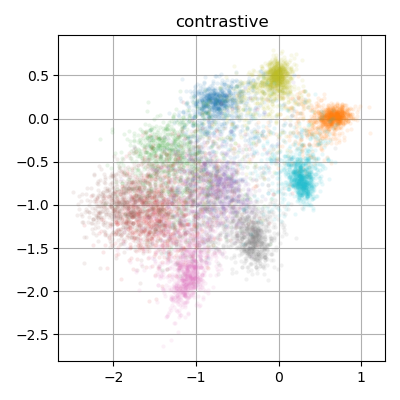}\includegraphics[scale=0.28]{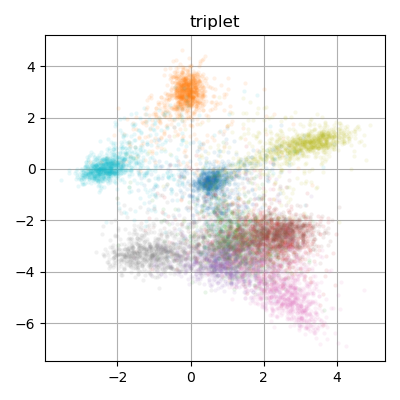}\includegraphics[scale=0.28]{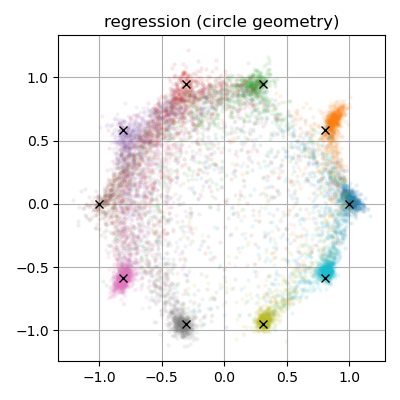}\includegraphics[scale=0.28]{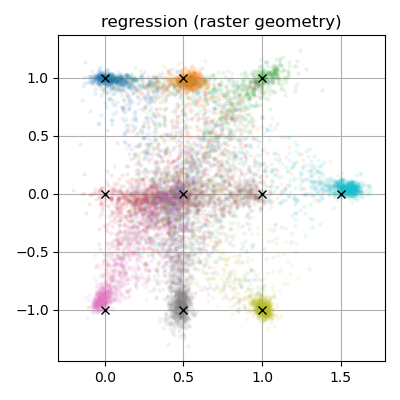}
\par\end{raggedright}
\bigskip{}

\begin{raggedright}
\includegraphics[angle=90,scale=0.27]{bilder/captions_links_fashion}\includegraphics[scale=0.28]{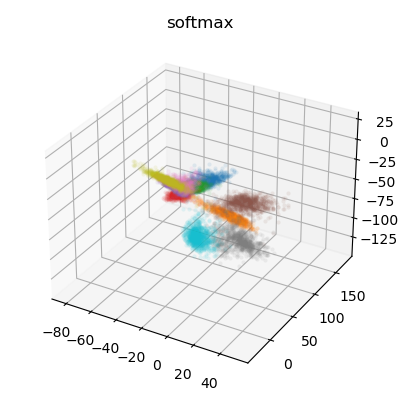}\includegraphics[scale=0.28]{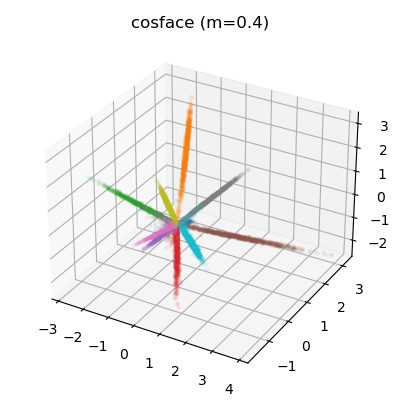}\includegraphics[scale=0.28]{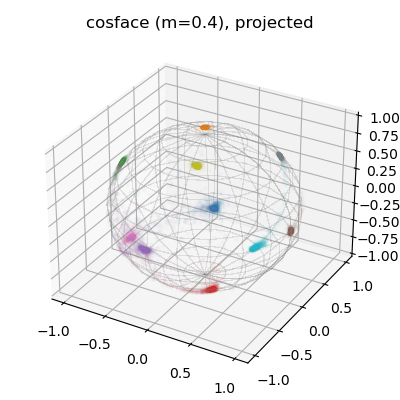}\includegraphics[scale=0.28]{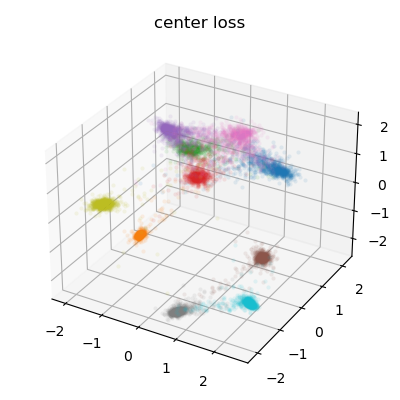}\includegraphics[scale=0.28]{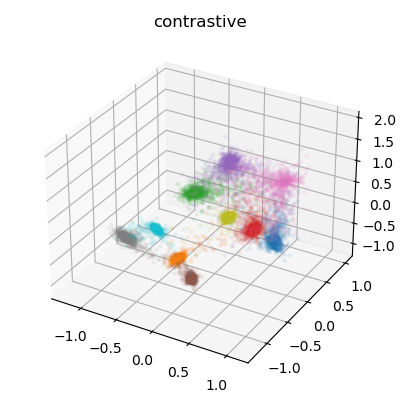}\includegraphics[scale=0.28]{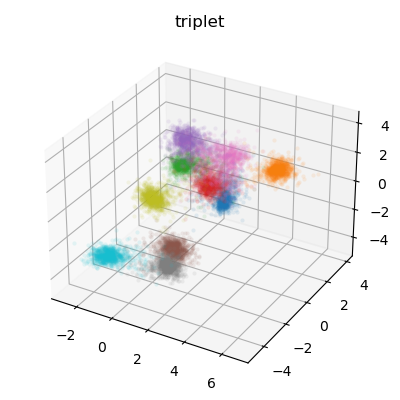}\includegraphics[scale=0.28]{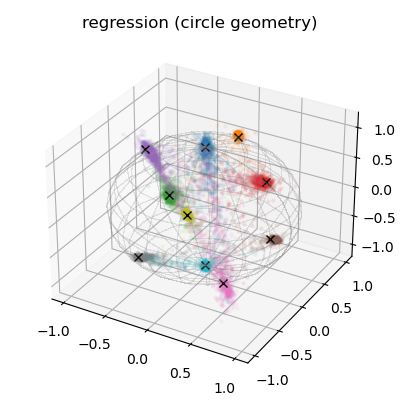}\includegraphics[scale=0.28]{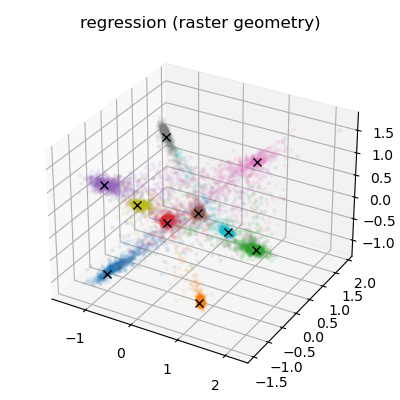}
\par\end{raggedright}
\begin{raggedright}
\includegraphics[angle=90,scale=0.27]{bilder/captions_links_cifar10}\includegraphics[scale=0.28]{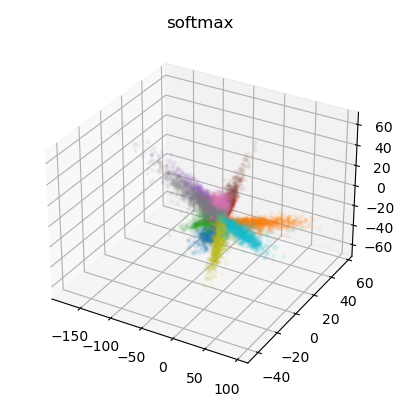}\includegraphics[scale=0.28]{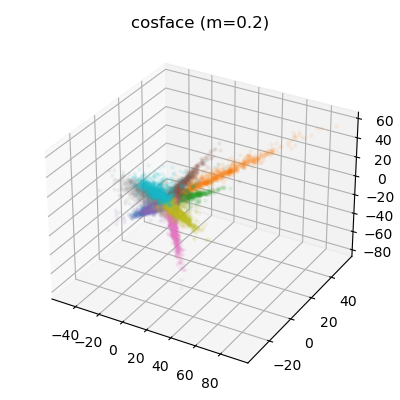}\includegraphics[scale=0.28]{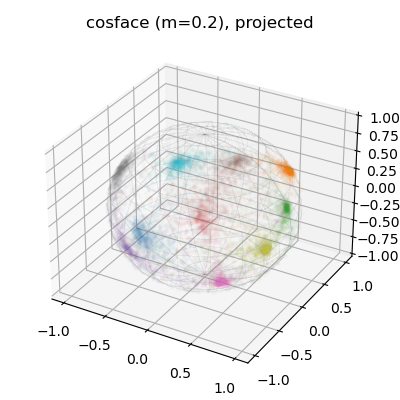}\includegraphics[scale=0.28]{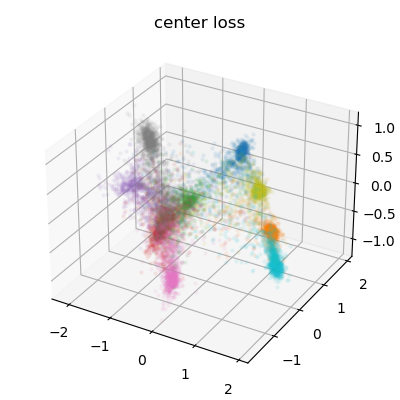}\includegraphics[scale=0.28]{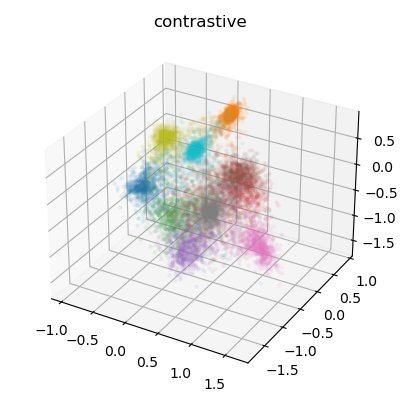}\includegraphics[scale=0.28]{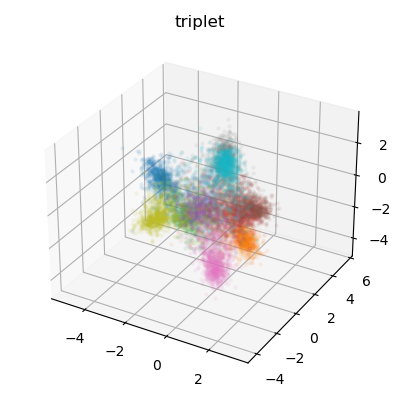}\includegraphics[scale=0.28]{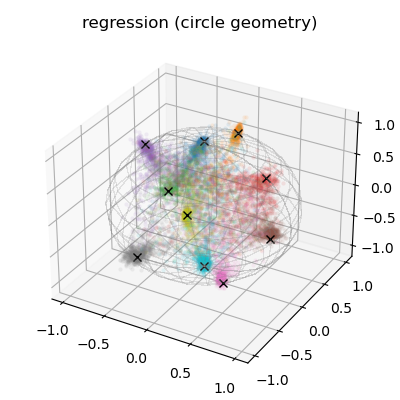}\includegraphics[scale=0.28]{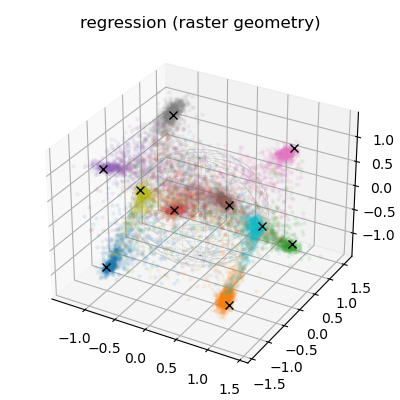}
\par\end{raggedright}
\caption{\label{fig:Comparison-1} 2-D and 3-D embeddings for MNIST, Fashion
MNIST and CIFAR-10 for visual comparison}
\end{sidewaysfigure*}

Figure \ref{fig:Comparison-1} shows an overview of all presented
techniques for shaping the embeddings for MNIST, Fashion MNIST and
CIFAR-10. Since Fashion MNIST and CIFAR-10 are harder datasets, good
two-dimensional embeddings are difficult to obtain. Therefore, also
3-D embeddings are provided, which are typically more powerful. Higher
dimensions (>3) often lead to even more powerful embeddings, but prevent
direct visual inspection.

\section{Summary}

This paper presented different approaches to influence the geometry
of embedded spaces. The different techniques are based on the softmax,
center and angular margin losses. Normalization of weight vectors
and features additionally enforces more regular embedded structures.
Regression techniques are an interesting approach to hand-crafted
embedded spaces. The different techniques have been evaluated qualitatively
using the MNIST, Fashion MNIST and CIFAR-10 datasets for two- and
three-dimensional embeddings. Proper design and understanding of the
embedded space before final classification can enhance the interpretability
and enable better open-set recognition and few-shot learning for deep
classification tasks. 

\bibliographystyle{plain}
\bibliography{literatur}

\end{document}